# Enrichment of the NLST and NSCLC-Radiomics computed tomography collections with AI-derived annotations

Deepa Krishnaswamy[1], Dennis Bontempi[2], Vamsi Krishna Thiriveedhi[1], Davide Punzo, David Clunie[3], Christopher P Bridge[4], Hugo JWL Aerts[2,5,6], Ron Kikinis[1], Andrey Fedorov[1]

[1]Brigham and Women's Hospital, Boston, MA, USA
[2]Artificial Intelligence in Medicine (AIM) Program, Mass General Brigham, Harvard Medical School, Boston, USA
[3]PixelMed Publishing, Bangor, PA, USA
[4]Department of Radiology, Massachusetts General Hospital, Boston, MA, USA
[5]Department of Radiation Oncology, Brigham and Women's Hospital, Dana-Farber Cancer Institute, Harvard Medical School, Boston, USA
[6]Radiology and Nuclear Medicine, CARIM & GROW, Maastricht University, Maastricht, The Netherlands

*corresponding author(s): Deepa Krishnaswamy (dkrishnaswamy@bwh.harvard.edu)*

# Abstract

Public imaging datasets are critical for the development and evaluation of automated tools in cancer imaging. Unfortunately, many do not include annotations or image-derived features, complicating their downstream analysis. Artificial intelligence-based annotation tools have been shown to achieve acceptable performance and thus can be used to automatically annotate large datasets. As part of the effort to enrich public data available within NCI Imaging Data Commons (IDC), here we introduce AI-generated annotations for two collections of computed tomography images of the chest, NSCLC-Radiomics, and the National Lung Screening Trial. Using publicly available AI algorithms we derived volumetric annotations of thoracic organs at risk, their corresponding radiomics features, and slice-level annotations of anatomical landmarks and regions. The resulting annotations are publicly available within IDC, where the DICOM format is used to harmonize the data and achieve FAIR principles. The annotations are accompanied by cloud-enabled notebooks demonstrating their use. This study reinforces the need for large, publicly accessible curated datasets and demonstrates how AI can be used to aid in cancer imaging.

# Background & Summary

National Cancer Institute (NCI) Imaging Data Commons (IDC)[1] contains publicly available cancer imaging, image-derived and image-related data, co-located with tools for exploration, visualization, and analysis. Public imaging data contributed by various initiatives, including those from The Cancer Imaging Archive (TCIA)[2], is ingested into this repository, allowing users to query metadata corresponding to images, annotations, and clinical attributes of the publicly available collections to define relevant cohorts, or subsets, of data. The IDC platform is based on the Google Cloud Platform (GCP), which enables the co-location of data with cloud-based tools for its exploration and analysis. Using tools from GCP, users can form a subset of data (a cohort) that is specific to the task at hand. Users also have the option of creating and using virtual machines to run computationally intensive jobs. Lastly, all analysis steps can be documented using Google Colaboratory python notebooks and shared with others.

Publicly available imaging datasets including the annotation of organs, lesions, and other regions of interest can aid in the development of imaging biomarkers, but unfortunately, many datasets suffer from the limited amount of annotations available. Using IDC, we chose to generate AI annotations for two collections: the Non-small Cell Lung Cancer (NSCLC) Radiomics dataset[3,4] and the National Lung Screening Trial (NLST) dataset[5,6]. The NSCLC-Radiomics collection contains labeled tumors, and only partially labeled organs of interest (combination of lung, esophagus, heart, and spinal cord). The NLST dataset, though widely used by many researchers[5,6] does not contain any image annotations.

In order to annotate the series, we make use of publicly available pre-trained deep learning models for automatically generating annotations. These annotations include volumetric segmentation of organs, the labeling of the region of the body scanned (e.g. chest and abdomen), and landmarks that capture the inferior to superior extent of a selection of organs and bones. The first pre-trained model used is the nnU-Net framework[7] for volumetric segmentation of thoracic organs. Since the collections we are analyzing concern lung cancer, we chose this model as it produces segmentations of the thoracic organs at risk (heart, aorta, trachea, and esophagus). These regions are routinely used during treatment planning, and could all be affected by the presence of lung cancer (and therefore used to develop new biomarkers or validate published ones). Multiple configurations of the nnU-Net framework (2D vs 3D, low vs high resolution, with and without test-time augmentation) were first applied to the NSCLC-Radiomics collection to evaluate its performance, where the best performing model configuration was chosen for NLST evaluation. To enhance information about bone and organ landmarks as well as the region of the body, the publicly available body part regression was employed[8]. The body part regression method[8] is an unsupervised method trained on a diverse set of CT data and produces a continuous score for each transverse slice in a 3D volume. These slice scores correspond to specific landmarks obtained from training data, and can then be used to infer the body part region.

In order to interoperability with the existing tools, harmonize representation of the annotations with that of the images being annotated, and towards implementing the findability, accessibility, interoperability, and reuse (FAIR) principles of data curation[9], we leverage the Digital Imaging and Communications in Medicine (DICOM)[10] standard. Our dataset is encoded using standard DICOM objects containing volumetric segmentations, slice-level annotations and segmentation-derived radiomics features. Furthermore, it is accompanied by the complete cloud-ready analysis workflow in the form of Google Colaboratory notebooks that can be used to recreate the dataset, and by the examples demonstrating how to query and visualize those standard objects, and how to convert them into alternative representations.

# Methods

The methods used to perform the preprocessing, analysis, and post-processing of the results are described in detail below. All of the analysis was performed in Google Colaboratory notebooks, making it easier to reproduce our analysis. Figure 1 gives a general overview of the study and describes the creation of DICOM objects and the sharing of data and code.

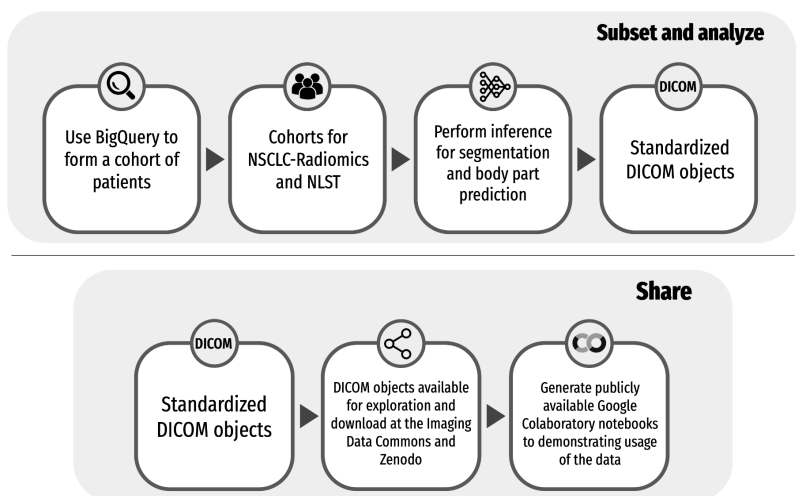

*Figure 1 - General overview of the study including steps needed to create the DICOM objects and how code and data are publicly shared.*

## Image collections analyzed

The NSCLC-Radiomics collection[3,4] is a radiology oncology dataset, where the patient population consisted of those with lung cancer. The dataset consists of data from 422 patients scanned at a single institution, as part of the study investigating whether radiomics features can be used to improve cancer detection[3,4].

The NLST collection[5,6] resulted from the clinical trial that investigated whether low-dose Computed Tomography imaging (CT) could be used to reduce the chance of mortality in a high-risk population of heavy smokers. NLST was a multicenter randomized controlled trial of non-contrast, non-ECG-gated low-dose chest CT for lung cancer screening, where participants were enrolled from 2002 to 2004 and scanned between 2002 and 2007. Participants were included in the study if they were between the ages of 55 and 74 and also smoked more than 30 packs per year, or had quit within the past 15 years. The collection consists of CT images in DICOM format for over 25,000 patients scanned at multiple time points, with a total of over 200,000 DICOM imaging series. Patients from 33 institutions were scanned using imaging equipment from a variety of vendors and utilizing a range of convolutional kernels.

## Selection of images for analysis

As a prerequisite for the analysis, each CT series must be assembled into a 3D volume. In order to enable such reconstruction, individual slices of the image series must possess consistent attributes that define its geometry, such as the pixel spacing, slice thickness and image orientation, and have no missing slices. Through use of the IDC platform and integration with GCP, it is possible to filter image series that cannot be reconstructed into a 3D volume using SQL queries interrogating DICOM image metadata, without having to download the image files.

The following criteria were applied to both the NSCLC-Radiomics collection and the NLST collection to select imaging series suitable for the analysis by the AI tools.
1. Consistent orientation of the slices within the series (based on the values of the DICOM ImageOrientationPatient attribute);
2. Consistent in-plane resolution (based on the DICOM PixelSpacing attribute);
3. No overlapping slices within the series (each value of the DICOM ImagePositionPatient within the series must be unique);
4. Consistent (within tolerance) spacing between the adjacent slices within the series (based on the difference in the values of ImagePositionPatient);
5. Series that were identified as localizer, or scout, scans based on metadata were excluded (based on the ImageType attribute).
6. A single series was chosen to be analyzed from each study of a selected patient (based on the first SeriesInstanceUID in the list).

Additional selection criteria were applied for the NLST collection:
1. Greater than 100 slices: Some patients may have incomplete scans, and therefore this criterion was introduced to detect potential issues.
2. SliceThickness greater than 1.5mm and less than 3mm: we observed the distribution of SliceThickness values for patients in the collection and imposed these criteria in order to remove potential outlier cases.
3. Only patients that screened positive for cancer were considered (this selection criterion was used to select a manageable size of the cohort, since in this study we did not aim to analyze the entire NLST collection due to its large size).

Version 10 of IDC data was used for the query. For further details, please refer to Appendix A.

After applying the selection criteria, cohorts of 414 patients (414 CT series) and 571 patients (1,039 CT series) were identified for analysis from the NSCLC-Radiomics and NLST collections, respectively. Please refer to Appendix A for an overview of the selection process for both collections in terms of the number of series. Table 1 includes information concerning the number of patients, studies, and series before and after filtering for the two collections. For a list of the series processed for NSCLC-Radiomics and NLST, please refer to https://github.com/ImagingDataCommons/nnU-Net-BPR-annotations/blob/main/common/queries/zenodo_nsclc_radiomics_series_analyzed.csv and

https://github.com/ImagingDataCommons/nnU-Net-BPR-annotations/blob/main/common/queries/zenodo_nlst_series_analyzed.csv respectively.

|  | NSCLC-Radiomics | | NLST | |
| --- | --- | --- | --- | --- |
|  | Entire collection | Analyzed subset | Entire cohort | Analyzed subset |
| Number of patients | 422 | 414 | 26,408 | 26,408 |
| Number of studies | 422 | 414 | 73,562 | 73,562 |
| Number of series | 422 | 414 | 204,319 | 204,319 |

*Table 1 - Characteristics for each analyzed collection before and after applying selection filters.*

For processing the cohort, the queries were run once. The Unique Resource Identifiers (URIs) corresponding to the files of the selected series were used to retrieve the files from the IDC Google Storage buckets.

## Preprocessing and processing steps

Figure 2 gives an overview of the preprocessing and processing steps in the pipeline. The DICOM files were downloaded from publicly hosted GCP buckets and sorted using the dicomsort package[11]. The package dcm2niix[12] was used to perform the conversion of DICOM to the NifTi format required by the analysis tools. Three of the series that passed the BigQuery checks had inconsistent attributes that were not considered in the query, leading to failure to convert into NIfTI format, and were subsequently discarded. The processing pipeline consists of two streams: 1) volumetric segmentation of the regions of interest; and 2) anatomic region and landmark annotation. Inference was performed using a pre-trained model for both use cases. Radiomics shape features were computed from the volumetric segmentations of the regions of interest. Finally, DICOM Segmentation and Structured Report objects were created for archival representation of the obtained analysis results.

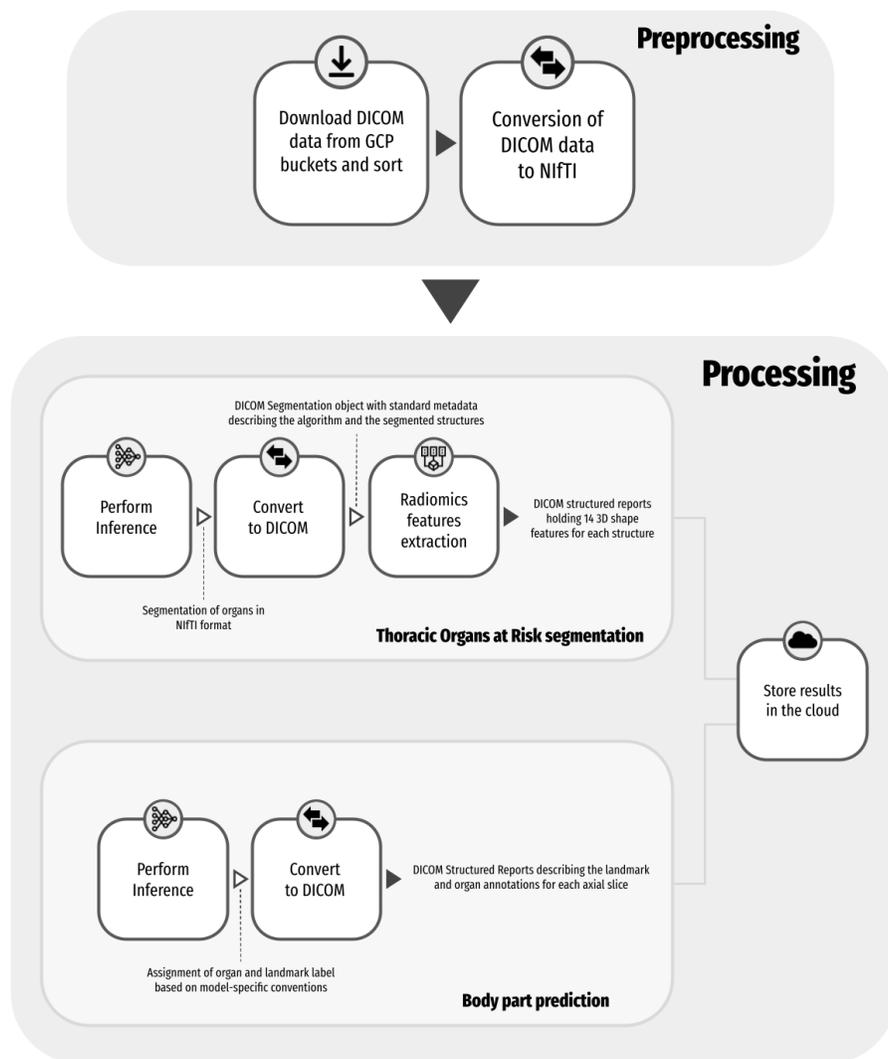

*Figure 2 - Preprocessing steps and processing steps for both segmentation of organs and body part landmark and region annotation*

## Anatomic region segmentation pipeline

### Inference using a pre-trained model

The nnU-Net deep learning framework[7] was used for automatic segmentation of the heart, aorta, trachea and the esophagus. The nnU-Net framework introduced a data-driven heuristic for hyperparameter selection that has proven to be very efficient for small datasets, yielding models that performed well in a number of challenges, including The Medical Segmentation Decathalon[13] and AMOS 2022[14]. These pretrained models, developed for different imaging modalities and various body regions[15] were shared with the original publication. nnU-Net performance and robustness in the aforementioned challenges made it a de-facto popular baseline for benchmarking other deep learning models. The Task055 nnU-Net model we used is part of the collection of pretrained models and was trained on data from the SegTHOR challenge[16], where the goal was to delineate thoracic organs of interest for the purposes of radiotherapy.

The authors of nnU-Net provided a python package[17] (v1) to run training, evaluation and inference. The implementation includes a command-line tool to easily run inference using a pre-trained model. The pre-trained model for Task055 SegTHOR was downloaded from Zenodo[18]. nnU-Net provides as output a single NifTi file containing the segmented structure (label 1 = esophagus, label 2 = heart, label 3 = trachea, label 4 = aorta).

**Radiomics feature extraction**

Pyradiomics package[19] (version v3.0.1) was employed to extract 3D shape features from the regions segmented by nnU-Net. The extracted features are provided in Appendix B. The pyradiomics package is an open-source library of functions required to extract a variety of radiomic features from a set of regions.

## Body part landmark and region annotation pipeline

**Inference using a pre-trained model**

The body part regression method[8] is an unsupervised method developed to provide slice level annotations of the presence of anatomical landmarks for axial CT volumes. The locations of the landmarks are then used to infer which parts of the body each axial slice corresponds to. There are numerous advantages to using such a method – the method does not rely on segmentations to train the neural network, thereby affording ease of use in case of retraining. The method produces not only annotations of organs identified in each axial slice, but also a set of landmarks that can assist in further downstream applications. The pretrained model performs localization of 35 bone and organ landmarks, including the cervical, thoracic and lumbar vertebrae. Additionally, six body part regions are defined from the landmarks including the head, shoulder-neck, chest, abdomen, pelvis and legs.

The authors provided a python package[20] to perform inference using a pre-trained model. The pretrained model was downloaded from Zenodo[23] (version 1.1). The output of the model provides a json file that holds the values of the slice scores, the corresponding landmark lookup table, and the slice indices that correspond to each body part region.

Figure 3 gives examples of the automatically generated annotations including the thoracic organ segmentations from nnU-Net, and the body part prediction landmarks and regions.

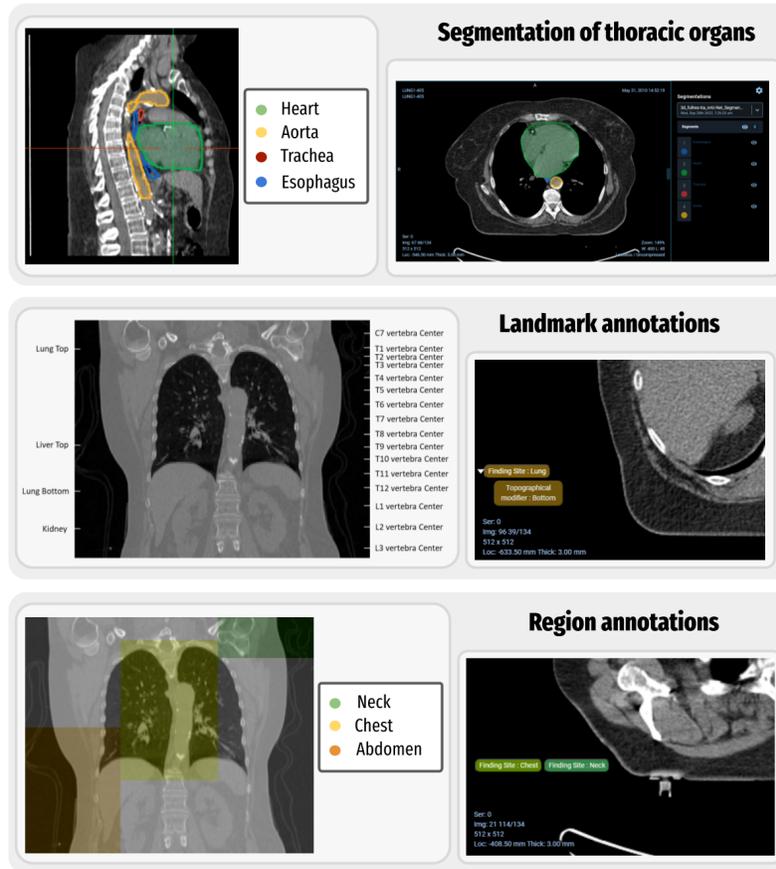

*Figure 3 - Examples of automatically generated annotations (top) nnU-Net segmentation examples showing a multi-planar reconstruction view (middle) landmark annotation example for a series (bottom) region annotation example for a series*

## Applications

There are numerous downstream applications in analyzing data from the NLST collection that may potentially benefit from the availability of anatomical region segmentations, radiomics extraction from those regions, the localization of bone and organ landmarks, as well as the labeling of regions:

- Segmentation of the anatomic structures is a common preprocessing step in image analysis pipelines (e.g., evaluation of cardiovascular risk using quantitative analysis of calcium relies on segmentation of the heart as a prerequisite[22], or quantification of airway obstruction/emphysema relies on the segmentation of airways including the trachea[23]).
- Segmentation of thoracic structures is necessary in the identification of organs-at-risk for radiotherapy treatment planning[16]
- Segmentations generated from the multiple nnU-Net models enable evaluation of the generalizability of the algorithm on an external dataset, and establish a baseline for evaluation of alternative similar algorithms. It becomes straightforward to evaluate the results produced by the nnU-Net segmentation, which can be considered state-of-the-art, and identify its weaknesses to motivate further improvement.
- Shape radiomics features can be used to detect potential outliers in the segmentations, but also can be utilized as quantitative features to stratify patients within the cohort or contribute to applications that utilize such features in more complex analyses.
- Annotations of landmarks and regions can be used to assist in defining regions of interest to simplify the task of segmentation of those structures.
- Given slice-level annotations of the landmarks and anatomic regions, it becomes possible to define anatomy-based search filters. As an example, one can select image volumes that contain specific vertebrae levels, or those that contain abdominal organs or head/neck, within a collection that focuses on chest imaging.

While the algorithms we used are publicly available for anyone to apply to the images we analyzed, development of the reproducible analysis workflows requires significant effort. Filtering of the images suitable for processing requires understanding of DICOM data representation and intricacies of developing selection queries. Inference pipelines computation takes time, bears costs in terms of resources used for the computation and, when using cloud resources, requires understanding of how cloud resources can be used cost- and time-efficiently. Therefore, we believe there is significant benefit in providing the resulting analysis artifacts to the community to enable aforementioned applications.

## Data Records

To ensure FAIR representation of data and metadata, all of the analysis results produced were converted into DICOM representation. DICOM standard is the international standard for medical images and related information[10]. It includes capabilities to describe, encode and exchange results of image analysis, such as those produced in the process of our study.

In total, four DICOM files were created for each series analyzed:
1. A Segmentation object that holds the nnU-Net predictions
2. A Structured Report that holds the radiomics shape feature computations
3. A Structured Report that holds the body part landmarks
4. A Structured Report that holds the body part regions

All of the data that are described here are available as part of the Zenodo "AI-derived annotations for the NLST and NSCLC-Radiomics computed tomography imaging collections" [18]
The data is also available within IDC as of version v13. As the data followed the standard DICOM representation, it was possible to store the data into a Google Healthcare DICOM datastore and automatically extract the metadata. This allows for the data to then be searchable and queried easily. Once stored in the DICOM datastore, it was then possible to visualize images and corresponding volumetric and slice-level annotations using the OHIF DICOM viewer[21] integrated with IDC[24]. We further demonstrate those features in the subsequent sections and the accompanying materials.

## Segmentations

Each label map was first saved as a separate NifTi file (.nii.gz). The "itkimage2segimage" tool from the software package dcmqi[25] was used to convert the segments from nnU-Net output to a DICOM Segmentation object. The annotations for the four thoracic organs for each of the analyzed CT series were saved as a single DICOM Segmentation object. Each SEG object contains a list of the ReferencedSOPInstanceUIDs within the ReferencedSeriesSequence that refer back to the original CT slices. Segment-level DICOM attributes document the type of the algorithm used, and coded terms describing the content of the segment (see Table 1 in Appendix C).

## Radiomics features

The 3D shape radiomics features were saved as Enhanced Structured Reports (SR) with the TID1500 template using the command "tid1500writer" from the dcmqi[25] package version 1.2.5. A single SR was saved for each series analyzed. Please refer to Table 2 in Appendix D for the list of features generated. Each feature is defined in the IBSI standard[26] and is described by the coded quantity and units.

## Body part regression landmarks

The landmarks produced from the body part prediction neural network were saved as Comprehensive 3DSR DICOM SRs, where a single report was saved for each series analyzed, and individual landmarks are associated with the image slices through the references to the corresponding SOPInstanceUIDs. The highdicom package[24] version 0.20.0 was used to generate the SR. Please refer to Table 3 in Appendix E for the metadata concerning the mapping of each of the body part regression codes to specific bones and landmarks.

## Body part regression regions

Inference from the body part prediction neural network also includes a region(s) assignment for each axial slice. This region assignment information was saved as Comprehensive 3DSR DICOM SRs, where a single report was saved for each series analyzed, and body part assignment was performed at the slice level via the referenced SOPInstanceUID. The highdicom package[27] version 0.20.0 was used to generate the SR. Please refer to Table 4 in Appendix F which holds the mapping of each body part region to specific target regions.

# Technical Validation

In the following sections we summarize the results of the evaluation. All of those results are available for interactive exploration using the Google Colaboratory notebooks accompanying the publication.

## Image preprocessing

The Methods section concerning the selection of images for analysis described the series of filters that were used to create a curated dataset for both the NSCLC-Radiomics and NLST collections. Results of the query were stored in a BigQuery table, to ensure the same series were used for subsequent steps of the analysis.

As described previously, the CT DICOM files were converted to the NifTi format, as this format was used as input for both the nnU-Net prediction as well as the body part regression method. The package dcm2niix[12] version 1.0.20220720 was used to perform the conversion. If multiple NifTi volumes were created from the conversion, the file with the number of slices matching the number of DICOM files was chosen. Further processing of the series was skipped if other errors were produced by dcm2niix.

## DICOM object validation

Before analyzing the results in the DICOM segmentation objects and structured reports, they were validated using publicly available tools. To check the DICOM conformance of the segmentation objects and structured reports, the tool *dciodvfy* from *dicom3tools*[28] was used. *dicom3tools* is a set of command line utilities that can create, validate and modify DICOM files, and also perform conversion to DICOM. For the SR objects, the software package *DICOMSRValidator* from *PixelMed*[29] was used to check conformance with the TID1500 DICOM standard SR template.

## Segmentation analysis

### Comparison of the NSCLC-Radiomics expert segmentations to the AI-derived segmentations

A subset of the series from the NSCLC-Radiomics collection contains expert segmentations of the heart and the esophagus. Therefore, it was possible to quantitatively assess the results by evaluating the Dice score and Hausdorff distance between the expert- and AI-derived segmentations. Figure 4 displays the Dice score metric of the heart for the three nnU-Net models evaluated, where in general high overlap results are demonstrated with all three AI-derived models. Interactive notebooks and plots accompanying this manuscript allow picking a point with the lowest Dice score for the 3D full resolution and test time augmentation model to visualize images and annotations using the OHIF viewer[21]. The differences between the expert and AI-derived annotations may be due to variations in the criteria used to delineate the heart between the expert and the data used for the pre-trained nnU-Net model. Segmentations for the NSCLC-Radiomics collection were provided by a single radiation oncologist using manual contouring on 2D slices[3] where for SegTHOR[16] manual annotations were provided by a radiation oncologist following the criteria recommended by the Radiation Therapy Oncology Group 2[16].

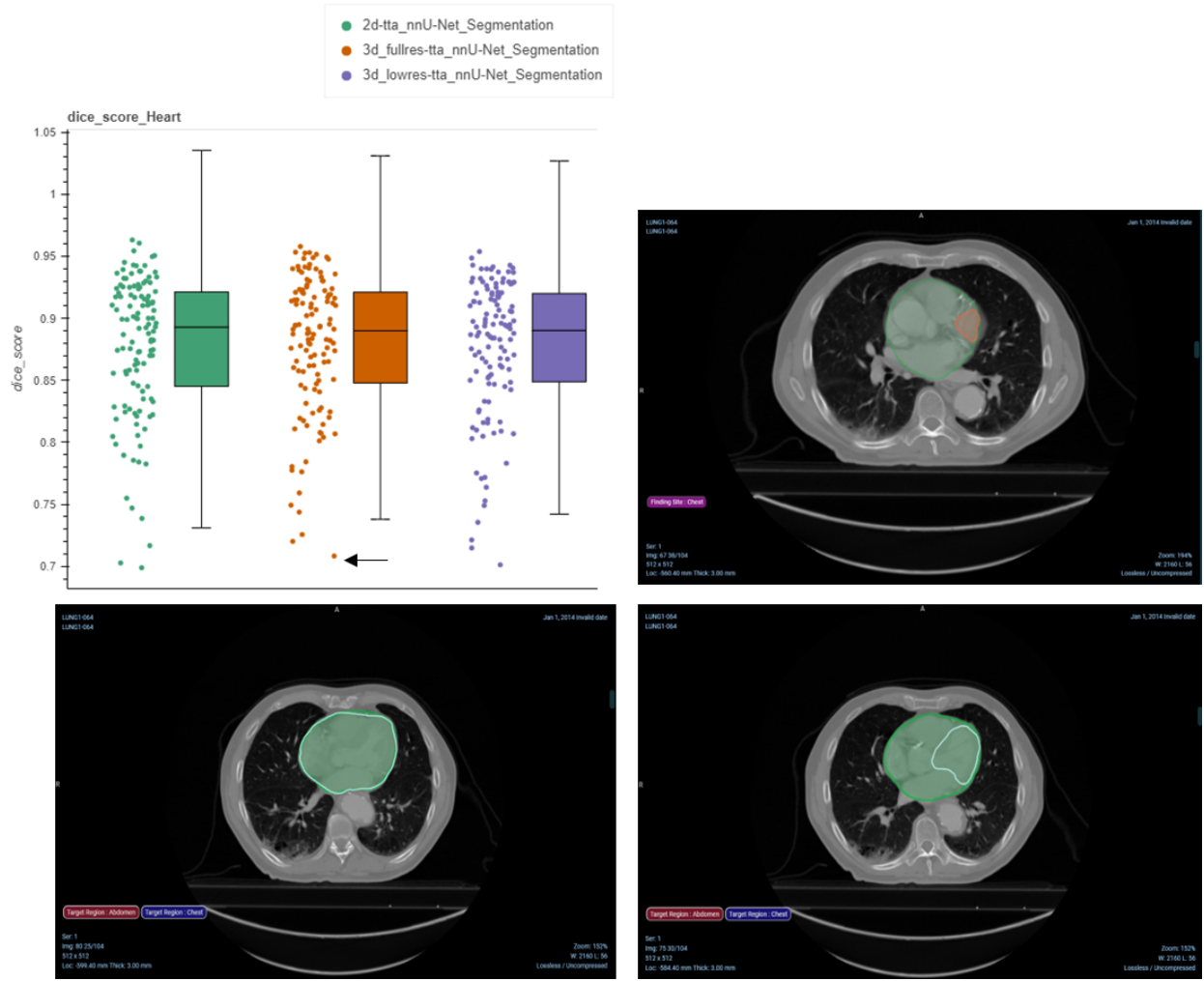

*Figure 4 - Evaluation of the AI-generated annotations with respect to the expert annotations of the heart for NSCLC-Radiomics. Top left: Dice score computation between each AI-generated annotation and the expert segmentation, Top right: point (highlighted by the black arrow) in the left pane corresponds to the visualization of the analysis results in the OHIF Viewer, with the ground truth in red and the AI-derived segmentation in green. Bottom left: Qualitative example of the nnU-Net model prediction in green to the ground truth in white with high overlap, Bottom right: Qualitative example of the nnU-Net model prediction in green to the ground truth in white with low overlap*

**Using shape radiomics features for outlier detection within AI segmentations for the NSCLC-Radiomics collection**

Another way to detect outliers in the segmentations apart from the Dice score and Hausdorff distance metrics is to perform a radiomics analysis, where features with high variance or with outliers may point to problematic analysis results. We compared the radiomics features extracted from the three AI-derived models to the features from the expert segmentations for NSCLC-Radiomics. Figure 5 displays this for the sphericity feature of the heart, where values closer to 1 indicate higher sphericity. We can see that the 3D full resolution model displays a narrow distribution of high sphericity values compared to the 3D low resolution and the 2D resolution models, which are expected as the 3D full resolution should ideally perform more accurately for our volumetric data. We can also see that the expert segmentations have a wide distribution, which may be due to inconsistencies with the delineation, as it was performed slice-wise. Picking the series with the lowest value of sphericity (left) yields the 2D MPR view in OHIF, where as expected, it can be seen that the 2D model does not perform well for the series.

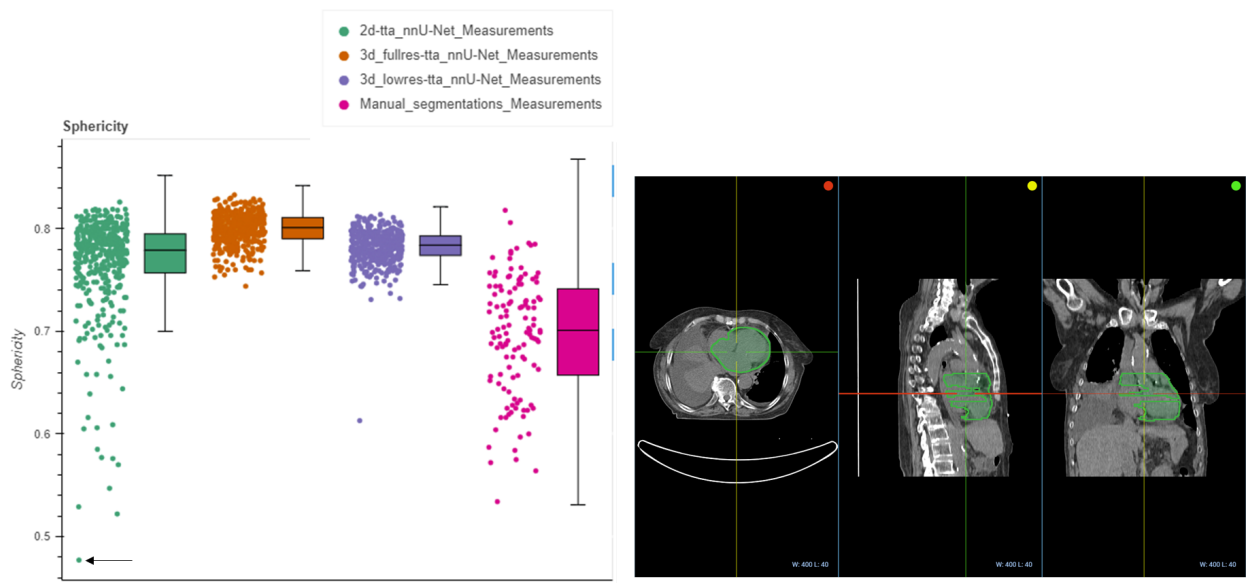

*Figure 5 - Evaluation of the heart sphericity radiomics features from the AI-generated annotations compared to the expert from NSCLC-Radiomics. Left: Radiomics distribution computed from each AI-generated annotation and expert annotation, and Right: point (highlighted by the black arrow) in the left pane corresponds to the visualization of the analysis results in the OHIF viewer, for the lowest sphericity value.*

**Using shape radiomics features for outlier detection within AI segmentations for the NLST collection**

Unlike the NSCLC-Radiomics collection, the NLST collection does not contain ground truth segmentations. Therefore, the analysis of the radiomics features for outlier detection was relied upon. Figure 6 below demonstrates the distributions of sphericity values for the four segmented regions. We can see that though in general the points are within the whisker ranges, some go beyond, for instance in the aorta. Picking the point with the lowest sphericity value for the aorta yields the case where it has not been properly segmented. This could be due either to the performance of the pre-trained model not being able to generalize to new data, or due to abnormalities in the patient.

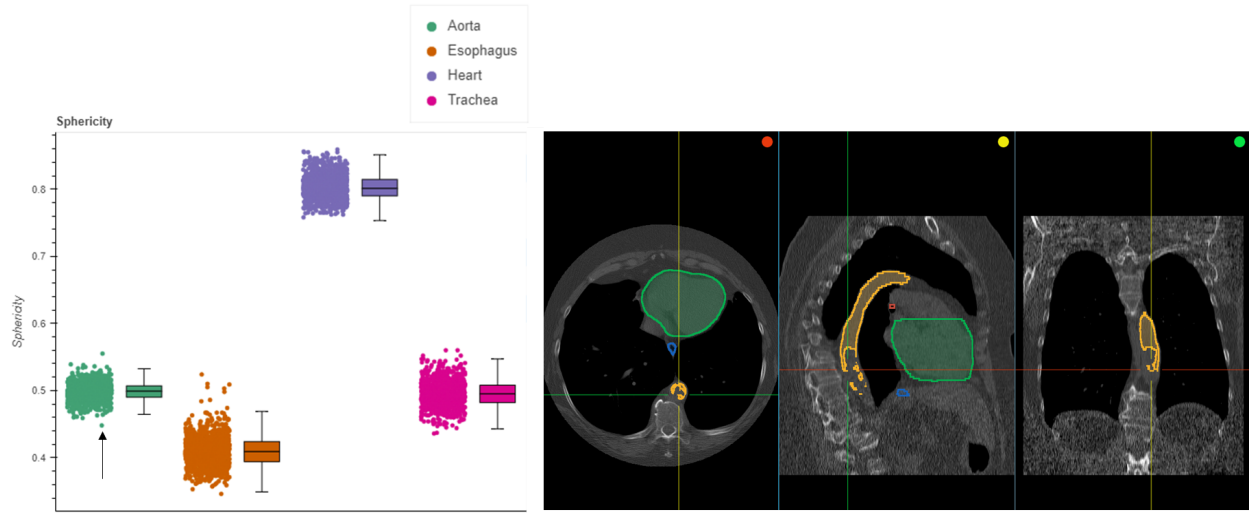

*Figure 6 - Evaluation of the sphericity radiomics features from the AI-generated annotations from NLST. Left: Radiomics distribution computed for the trachea, esophagus, heart and aorta, and Right: point (highlighted by the black arrow) in the left pane corresponds to the visualization of the analysis results in the OHIF viewer, for the lowest value of the aorta.*

## Landmark analysis

**Validation of the lung start and end landmarks using the ground truth lung segmentation for NSCLC-Radiomics**

To assess the accuracy of the body part predicted landmarks, we evaluate their locations against the expert segmentations. For NSCLC-Radiomics, the segmentation of the left and right lungs is provided for a majority of the patients. Therefore we compared the bottom of the lung and top of the lung landmarks to the inferior and superior axial slice locations of the expert segmentations. Figure 7 (A) displays the distributions of these two sets of differences, where the 'lung bottom difference' represents the difference between the bottom of the lung landmark and the inferior axial slice location of the expert segmentation, and the 'lung top difference' represents the difference between the top of the lung landmark and the superior axial slice location of the expert. Figure 7 (B) demonstrates an example disagreement between the body part predicted slice and the expert, where the predicted landmark is more inferior. This could be due to abnormalities in the anatomy of the patient.

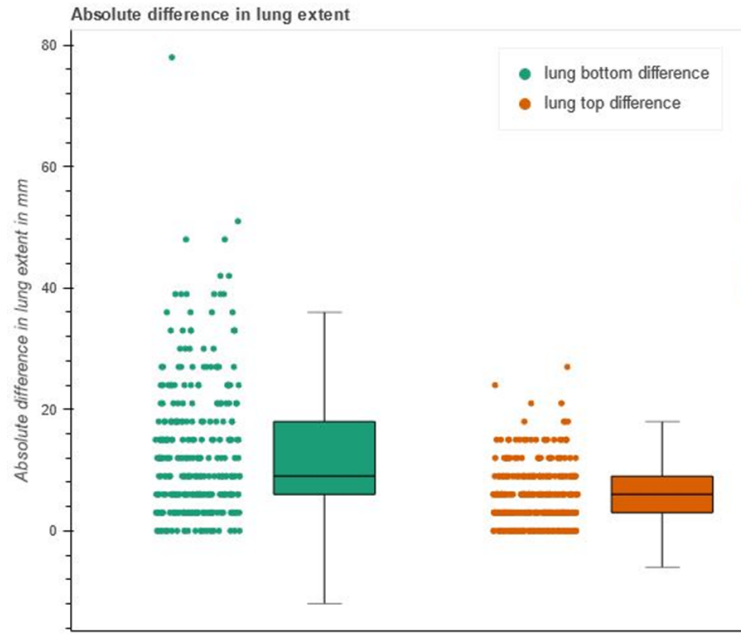

*(a) Difference between the expert lung segmentation and the BPR derived lung_start and lung_end landmarks.*

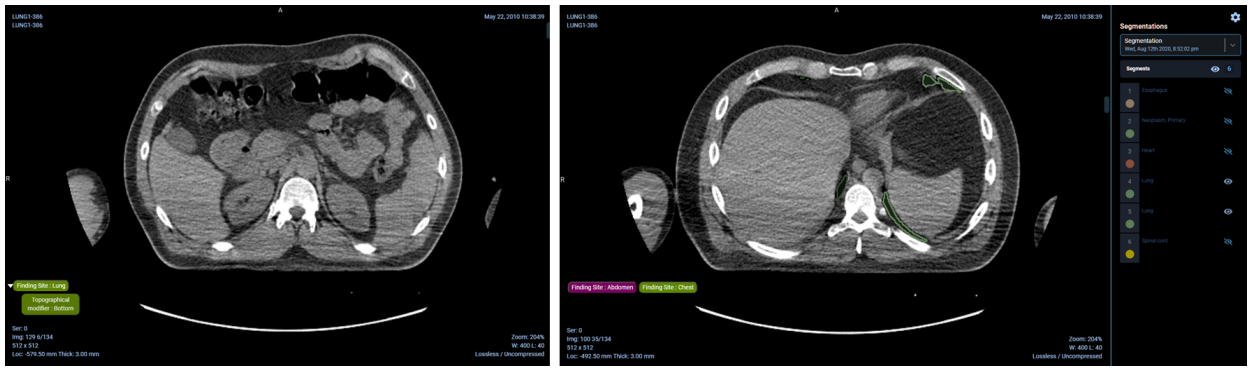

*(b) Left - Bottom of the lung landmark slice predicted by body part regression network, Right - Most inferior slice of the lung (green) according to the expert annotation*

*Figure 7 - Demonstration of the difference between the body part regression predicted landmarks and the expert lung segmentations (a) Boxplot displaying the difference between the most inferior point of the ground truth lung to the lung_start landmark, and the difference between the most superior point of the ground truth lung to the lung_end landmark and (b) Qualitative results in the OHIF viewer displaying the difference between the predicted landmark axial slice and the most inferior slice of the expert segmentation*

**Calculation of the lung extent**

In the absence of manual annotations for the NLST collections, we resort to the analysis of the distributions of various measures we can calculate from the produced annotations. One such measure is lung height, which we calculated the distance between the *lung_start* and *lung_end* landmarks in mm which correspond to the most inferior and most superior axial slices of the lung respectively. Figure 8 summarizes these height values for the NLST collection, where it can be seen that the median heights are relatively close to the expected height of the lungs in adults, which is 27 cm[30]. Lung heights that are significantly different from the median could indicate an issue with scanning the patient, for instance an incomplete scan, or could indicate an anatomical problem with the lung itself. Figure 8 shows an example of where the identification of the start slice of the lung is slightly off.

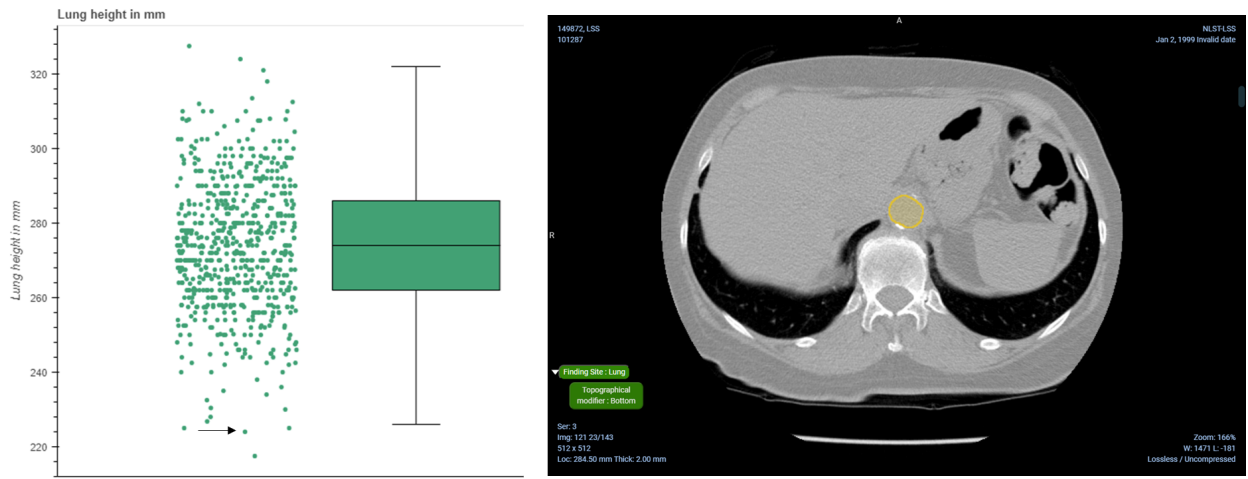

*Figure 8 - Evaluation of the distribution of distances between the start and end of the lungs in mm for the NLST collection. Left: Distribution of distances between the start and end of the lungs, and Right: point (highlighted by the black arrow) in the left pane corresponds to the visualization of the analysis results in the OHIF viewer, where the identification of the lung start slice is slightly off.*

## Anatomic region analysis

**Calculation of the ratio of slices assigned to each body part region**

The body part prediction method provides an assignment of regions to each axial slice, where more than one region may be assigned to a slice. To quantitatively assess assignment of these regions and summarize it over the entire collection, we plotted the ratio of the slices assigned to each body part region. Using this plot one could ensure that the largest proportion of slices was assigned to the chest region, which is expected, since the collections are focused on patients with lung cancer. Figure 9 displays these slice ratios for the NLST collection, where it can be seen that areas outside of the chest are scanned. We can visually confirm the presence of a portion of the head in the scan.

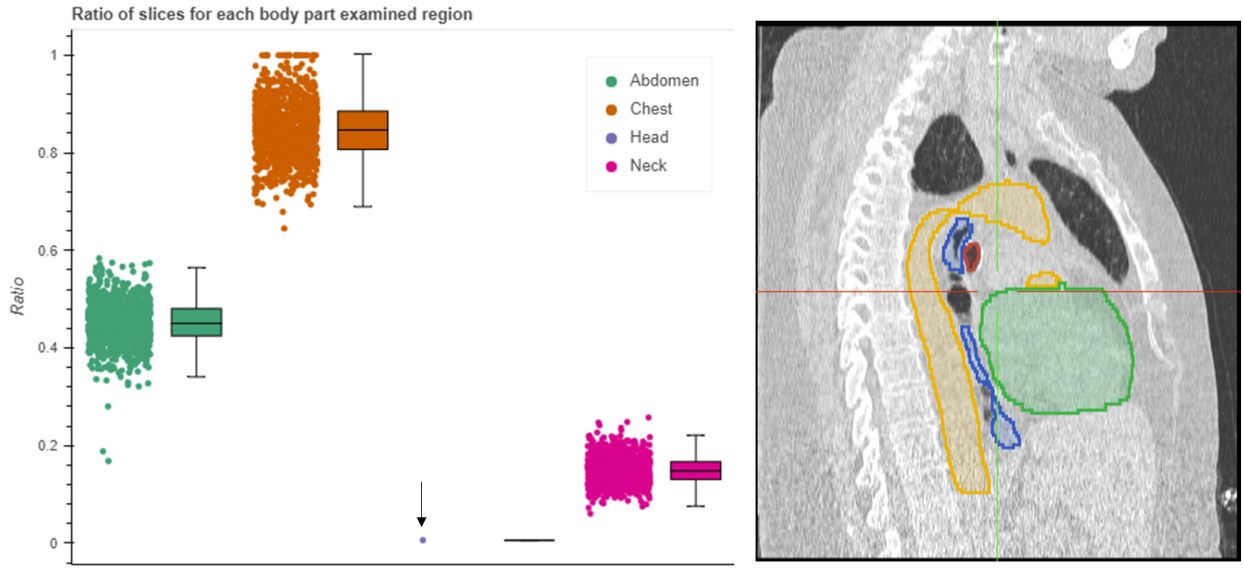

*Figure 9 - Evaluation of the percentage of slices assigned to each region (head, neck, chest, abdomen, pelvis and legs) for the NLST collection. Left: Distribution of the percentages for each region, and Right: point (highlighted by the black arrow) in the left pane corresponds to the visualization of the analysis results in the OHIF viewer, where we can confirm the scan includes a portion of the head.*

## Usage Notes

The annotations that we provide as DICOM Segmentation objects and Structured reports are ingested and publicly available in IDC. Users can visualize those annotations, build cohorts using the AI-derived annotations, download the objects, query for specific metadata of interest, perform analysis, and interact with the objects and data. Using the OHIF viewer (version 4.12.50), users have the ability to view the associated DICOM Segmentation objects and the DICOM Structured reports for the body part landmarks and regions. While most of the annotations are searchable using the IDC Portal, the shape features can only be queried using BigQuery SQL.

We have provided a Google Colaboratory notebook demonstrating the interaction with data for both the nnU-Net prediction analysis (including radiomics analysis), along with the body part prediction results analysis: https://github.com/ImagingDataCommons/nnU-Net-BPR-annotations/blob/main/usage_notebooks/scientific_data_paper_usage_notes.ipynb. To perform an additional exploratory analysis of the enriched collections, the Google LookerStudio dashboard can be utilized: https://datastudio.google.com/reporting/6bcdc67b-6d09-41da-b6d2-7e3102299347.

For the nnU-Net prediction segmentation analysis, we divide the notebook into the following sections:

1. *Examples of interactive plots comparing expert segmentations vs AI-derived segmentations*

   The NSCLC-Radiomics collection contains series that have some annotated organs (heart and lung), while NLST does not. To therefore choose the best pre-trained nnU-Net model, we chose three models and compared performance to the expert annotations. We then chose the most robust and high performing model and used that for the NLST collection. Therefore to prove our choice of model, for the NSCLC-Radiomics series with expert annotations, we quantitatively compared the performance of the three nnU-Net models to the expert delineations in terms of Dice and Hausdorff distance metrics. The user can interact with the plot and open links to the OHIF viewer.

2. *Examples of how to query and download DICOM Segmentation objects and the associated CT files*

   Users may want to understand how to query for DICOM Segmentation objects, to perhaps load in 3DSlicer, other external programs, or use locally. We therefore provide code with concrete examples of how to use BigQuery for querying for appropriate data and how to download the files from the public IDC buckets.

3. *Demonstration of how to convert DICOM files, for both the CT files and the segmentation objects using multiple packages*

   As many AI and ML pipelines require numpy arrays or NifTi files for holding the segmentation output, we demonstrate how to convert the DICOM Segmentation objects to NifTi. We use three robust software packages, dcmqi[25], pydicom_seg[31] and highdicom[27] to demonstrate multiple ways of performing this conversion. Dcmqi for instance converts each segment to a separate file (.nii.gz) and outputs a JSON file that holds the metadata of the DICOM Segmentation object, in particular the CodingSchemeDesignators, CodeValues and CodeMeanings necessary to interpret the anatomical object. 3DSlicer can be used to view the DICOM Segmentation objects, which internally uses dcmqi for the conversion to labelmaps. We also show how to convert CT DICOM files to NifTi using dcm2niix[12].

4. *Visualization of DICOM and NifTi files using ITKWidgets and custom code*

   Instead of downloading the DICOM and NifTi files (both CT files and the segmentation overlays) and viewing them in external programs, you may want to quickly view them within the notebook itself. We make use of ITKWidgets[32] and custom code for these demonstrations.

For the nnU-Net radiomics feature evaluation, we demonstrate the following:

1. *Examples of how to query using the radiomics features*

   The radiomics features are stored in the quantitative_measurements publicly accessible table. We show how querying the metadata is beneficial, instead of manually reading in a set of DICOM SRs and extracting the nested metadata yourself. We demonstrate how to extract radiomics features for a single series, or over all series in the collection, and also how to obtain a list of series that fall within a certain range of values.

2. *Examples of interactive plots with the radiomics features values*

   We demonstrate an interactive plot showing a single feature for all series and a single region for NSCLC-Radiomics, and a single feature for all series across all regions for NLST. These types of plots can be used to identify possible outliers in the segmentation. The user can also click on sphericity feature values which opens up an OHIF viewer link to examine specific series.

For the Body Part Regression analysis of landmarks, we demonstrate the following:

1. *Plot the location of the landmarks on a coronal slice*

   Though the landmarks per transverse slice can be visualized in OHIF, it is useful to get a quick assessment of the landmark locations for a patient. We therefore demonstrate the automatic ability to assess the transverse locations of the landmarks on a coronal slice (as demonstrated in Figure 3 of this paper).

2. *Evaluation of the lung landmark compared to the expert segmentations*

   Validation of the landmarks in a quantitative manner and not only qualitative is crucial, especially for the NSCLC-Radiomics dataset where expert lung segmentations are available. Therefore we compute the distance between the lung_top landmark with the superior point of the expert lung, and the lung_bottom landmark with the inferior point of the expert lung segmentation. The user can also interact with this plot and assess if there are any series where the body part prediction is incorrect.

3. *Evaluation of the lung landmark distribution*

   As the NLST collection does not have expert segmentations, one method to assess the body part prediction is to calculate the distance between the top and bottom lung landmarks and assess for outliers. We demonstrate this by creating an interactive plot again.

4. *Extract series that have specific landmarks*

   We demonstrate the usefulness of having the metadata extracted from the landmarks SRs in a table, as we can infer which patients do not have certain expected landmarks (aka lung).

5. *Download and extract values from the SRs*

   Though the metadata from the landmarks can be made available in a table, the user still has the option of downloading the files and extracting the values themselves. We demonstrate how to extract relevant fields using dcmqi[25] and highdicom[27].

For the Body Part Regression analysis of regions, we demonstrate the following use cases:

1. *Qualitative evaluation of the regions*

   Though one can view the body part assigned regions in OHIF, one may want a quick overview of the transverse slices assigned to each region. We demonstrate how to do this using matplotlib.

2. *Creation of an interactive plot for series that include the chest*

    We may be interested to know the percentage of slices that are assigned to the chest region. By analyzing these range of values we can see the extent of a scan and detect any outliers.

3. *Creation of an interactive plot for percentages of regions for all series*

    We may also be interested in quickly assessing differences in scan regions, by seeing if patients have the head, neck, pelvis or legs included. For instance, we may have included patients with brain scans, and would want to filter those scans from our analysis.

4. *Download and extract values from the SRs*

    Though the metadata from the regions can be made available in a table, the user still has the option of downloading the files and extracting the values themselves. We demonstrate how to extract relevant fields using dcmqi[25] and highdicom[27].

# Data Availability

The DICOM structured reports are available in IDC v13 at the following link https://portal.imaging.datacommons.cancer.gov/explore/filters/?analysis_results_id=nnU-Net-BPR-annotations. For the NSCLC-Radiomics collection, the following are included for each of the cases analyzed:
1. Three Segmentation objects that holds the nnU-Net predictions for 2d+tta, 3d_lowres+tta, and 3d_highres+tta, respectively
2. Three Structured Reports that hold the radiomics shape feature computations for 2d+tta, 3d_lowres+tta, and 3d_highres+tta, respectively
3. A Structured Report that holds the body part landmarks
4. A Structured Report that holds the body part regions

For the NLST collection, the following are included:
1. A Segmentation object that holds the nnU-Net predictions for 3d_highres+tta
2. A Structured Report that holds the radiomics shape feature computations for 3d_highres+tta
3. A Structured Report that holds the body part landmarks
4. A Structured Report that holds the body part regions

# Code Availability

The code for creating the annotations and demonstrating interactions with the data is Release v1.0.0: https://github.com/ImagingDataCommons/nnU-Net-BPR-annotations/releases/tag/v1.0.0, with the Colaboratory usage notebook available here: https://github.com/ImagingDataCommons/nnU-Net-BPR-annotations/tree/main/usage_notebooks. In order to run the notebook, users must set up a GCP project by following the instructions here https://learn.canceridc.dev/introduction/getting-started-with-gcp.

The notebooks used to query, run the analysis and create the DICOM Segmentation objects and Structured Reports are listed here:
1. NSCLC-Radiomics analysis:
   https://github.com/ImagingDataCommons/nnU-Net-BPR-annotations/blob/main/nnunet/notebooks/idc_nsclc_nnunet_and_bpr_infer.ipynb
2. NLST analysis:
   https://github.com/ImagingDataCommons/nnU-Net-BPR-annotations/blob/main/nnunet/notebooks/idc_nlst_nnunet_and_bpr_infer.ipynb

## Acknowledgements

This work has been funded in whole or in part with Federal funds from the National Cancer Institute, National Institutes of Health, under Task Order No. HHSN26110071 under Contract No. HHSN261201500003l. This work has additionally been funded in part from the T32 Image Guidance, Precision Diagnosis and Therapy Research Fellowship T32EB025823-04.

## Author contributions

Deepa Krishnaswamy was responsible for the body part prediction analysis and the generation of all DICOM structured report objects. Deepa was responsible for the qualitative and quantitative evaluation and the main contributor in terms of writing, preparing text and figures and editing of this manuscript. Dennis Bontempi was responsible for the nnU-Net analysis and the generation of the DICOM segmentation objects, and the overall organization of the code. Dennis was also a main contributor to the editing of this manuscript. Vamsi Thiriveedhi was responsible for the development of the queries for the body part regression landmarks and regions analysis. Davide Punzo was responsible for the OHIF viewer: parsing SR annotation, slice level annotation and segmentation display features. David Clunie was responsible for the correct encoding of the DICOM Segmentation objects and Structured Reports. Christopher Bridge assisted with the conversion of DICOM objects to other formats for anlaysis. Hugo Aerts was responsible for the formulation and editing of the manuscript. Ron Kikinis was responsible for the formulation and editing of the manuscript. Andrey Fedorov was responsible for the overall formulation and structure of the manuscript, reviewing of code, and extensive editing.

## Competing interests

There are no conflicts of interest.

# Appendices

## Appendix A

### NSCLC-Radiomics query

The query for performing the filtering of relevant series for the NSCLC-Radiomics collection is below. The query is also available here https://github.com/ImagingDataCommons/nnU-Net-BPR-annotations/blob/main/common/queries/NSCLC_Radiomics_query.txt.

```sql
WITH
  nsclc_radiomics_instances_per_series AS (
  SELECT
    DISTINCT(SeriesInstanceUID),
    COUNT(DISTINCT(SOPInstanceUID)) AS num_instances,
    COUNT(DISTINCT(ARRAY_TO_STRING(ImagePositionPatient,"/"))) AS position_count,
    COUNT(DISTINCT(ARRAY_TO_STRING(PixelSpacing,"/"))) AS pixel_spacing_count,
    COUNT(DISTINCT(ARRAY_TO_STRING(ImageOrientationPatient,"/"))) AS orientation_count,
    STRING_AGG(DISTINCT(SAFE_CAST("LOCALIZER" IN UNNEST(ImageType) AS string)),"") AS has_localizer
  FROM
    `bigquery-public-data.idc_current.dicom_all`
  WHERE
    collection_id = "nsclc_radiomics"
    AND Modality = "CT"
  GROUP BY
    SeriesInstanceUID
    ),

  distinct_slice_location_difference_values AS (
  SELECT
    -- SAFE_CAST(ImagePositionPatient[SAFE_OFFSET(2)] AS NUMERIC) - LAG(SAFE_CAST(ImagePositionPatient[SAFE_OFFSET(2)] AS NUMERIC),1) OVER(partition by SeriesInstanceUID ORDER BY SAFE_CAST(ImagePositionPatient[SAFE_OFFSET(2)] AS NUMERIC) DESC) AS SliceLocation_difference,
        DISTINCT(TRUNC(SAFE_CAST(ImagePositionPatient[SAFE_OFFSET(2)] AS NUMERIC),1) - LAG(TRUNC(SAFE_CAST(ImagePositionPatient[SAFE_OFFSET(2)] AS NUMERIC),1),1) OVER(partition by SeriesInstanceUID ORDER BY TRUNC(SAFE_CAST(ImagePositionPatient[SAFE_OFFSET(2)] AS NUMERIC),1) DESC)) AS SliceLocation_difference,
    SeriesInstanceUID,
    StudyInstanceUID
```

```sql
  FROM
    `bigquery-public-data.idc_current.dicom_all`
  ),

  nsclc_radiomics_values_per_series AS (
  SELECT
    COUNT(DISTINCT(distinct_slice_location_difference_values.SliceLocation_difference)) as num_differences,
    ANY_VALUE(distinct_slice_location_difference_values.StudyInstanceUID) AS StudyInstanceUID,
    distinct_slice_location_difference_values.SeriesInstanceUID AS SeriesInstanceUID,
    ANY_VALUE(nsclc_radiomics_instances_per_series.num_instances) AS num_instances,
    ANY_VALUE(CONCAT("https://viewer.imaging.datacommons.cancer.gov/viewer/",distinct_slice_location_difference_values.StudyInstanceUID,"?seriesInstanceUID=",distinct_slice_location_difference_values.SeriesInstanceUID)) AS idc_url,
    MAX(ABS(distinct_slice_location_difference_values.SliceLocation_difference)) as max_difference,
    MIN(ABS(distinct_slice_location_difference_values.SliceLocation_difference)) as min_difference
  FROM
     distinct_slice_location_difference_values
  JOIN
    nsclc_radiomics_instances_per_series
  ON
    nsclc_radiomics_instances_per_series.SeriesInstanceUID = distinct_slice_location_difference_values.SeriesInstanceUID
  WHERE
    nsclc_radiomics_instances_per_series.num_instances/nsclc_radiomics_instances_per_series.position_count = 1
    AND nsclc_radiomics_instances_per_series.pixel_spacing_count = 1
    AND nsclc_radiomics_instances_per_series.orientation_count = 1
    AND has_localizer = "false"
  GROUP BY
    distinct_slice_location_difference_values.SeriesInstanceUID)

  SELECT
    dicom_all.PatientID,
    dicom_all.StudyInstanceUID,
    dicom_all.SeriesInstanceUID,
    dicom_all.Modality,
    nsclc_radiomics_values_per_series.num_instances,
    nsclc_radiomics_values_per_series.num_differences,
    nsclc_radiomics_values_per_series.max_difference,
```

```
    nsclc_radiomics_values_per_series.min_difference,
    nsclc_radiomics_values_per_series.idc_url,
    dicom_all.gcs_url
  FROM
    `bigquery-public-data.idc_current.dicom_all` AS dicom_all
  JOIN
    nsclc_radiomics_values_per_series
  ON
    dicom_all.SeriesInstanceUID = nsclc_radiomics_values_per_series.SeriesInstanceUID
  WHERE
    # nsclc_radiomics_values_per_series.num_differences = 1
    nsclc_radiomics_values_per_series.num_differences <= 2
    AND
nsclc_radiomics_values_per_series.max_difference/nsclc_radiomics_values_per_series.min_difference < 2
  ORDER BY
    dicom_all.PatientID
```

## NLST query

The query for performing the filtering of relevant series for the NLST collection is below. The query is also available here https://github.com/ImagingDataCommons/nnU-Net-BPR-annotations/blob/main/common/queries/NLST_query.txt.

```
WITH
  nlst_positive AS (
  SELECT
    SAFE_CAST(pid AS STRING) AS pid,
    PatientID,
    StudyInstanceUID,
    SeriesInstanceUID
  FROM
    `bigquery-public-data.idc_current.nlst_prsn` AS nlst
  JOIN
    `bigquery-public-data.idc_current.dicom_all` AS dicom_all
  ON
    SAFE_CAST(nlst.pid AS STRING) = dicom_all.PatientID
  WHERE
    can_scr = 1
```

```sql
  ),

  nlst_instances_per_series AS (
  SELECT
    ANY_VALUE(nlst_positive.PatientID) AS PatientID,
    nlst_positive.StudyInstanceUID,
    nlst_positive.SeriesInstanceUID,
    COUNT(DISTINCT(SOPInstanceUID)) AS num_instances,
    COUNT(DISTINCT(ARRAY_TO_STRING(ImagePositionPatient,"/"))) AS position_count,
    COUNT(DISTINCT(ARRAY_TO_STRING(PixelSpacing,"/"))) AS pixel_spacing_count,
    COUNT(DISTINCT(ARRAY_TO_STRING(ImageOrientationPatient,"/"))) AS orientation_count,
    MIN(SAFE_CAST(SliceThickness AS float64)) AS min_SliceThickness,
    MAX(SAFE_CAST(SliceThickness AS float64)) AS max_SliceThickness,
    MIN(SAFE_CAST(ImagePositionPatient[SAFE_OFFSET(2)] AS float64)) AS min_SliceLocation,
    MAX(SAFE_CAST(ImagePositionPatient[SAFE_OFFSET(2)] AS float64)) AS max_SliceLocation,
    STRING_AGG(DISTINCT(SAFE_CAST("LOCALIZER" IN UNNEST(ImageType) AS string)),"") AS has_localizer,
    ANY_VALUE(dicom_all.ImageOrientationPatient) AS ImageOrientationPatient,
    ANY_VALUE(dicom_all.Modality) AS Modality
  FROM
    `bigquery-public-data.idc_current.dicom_all` as dicom_all
  JOIN
    nlst_positive
  ON
    nlst_positive.SeriesInstanceUID = dicom_all.SeriesInstanceUID
  WHERE
    collection_id = "nlst"
    AND Modality = "CT"
  GROUP BY
    StudyInstanceUID,
    SeriesInstanceUID ),

  distinct_slice_location_difference_values AS (
  SELECT
      # DISTINCT(SAFE_CAST(ImagePositionPatient[SAFE_OFFSET(2)] AS NUMERIC) - LAG(SAFE_CAST(ImagePositionPatient[SAFE_OFFSET(2)] AS NUMERIC),1) OVER(partition by SeriesInstanceUID ORDER BY SAFE_CAST(ImagePositionPatient[SAFE_OFFSET(2)] AS NUMERIC) DESC)) AS SliceLocation_difference,
      DISTINCT(TRUNC(SAFE_CAST(ImagePositionPatient[SAFE_OFFSET(2)] AS NUMERIC),1) - LAG(TRUNC(SAFE_CAST(ImagePositionPatient[SAFE_OFFSET(2)] AS NUMERIC),1),1) OVER(partition by SeriesInstanceUID ORDER BY TRUNC(SAFE_CAST(ImagePositionPatient[SAFE_OFFSET(2)] AS NUMERIC),1)
```

```sql
  DESC)) AS SliceLocation_difference,
      SeriesInstanceUID,
      StudyInstanceUID
  FROM
      `bigquery-public-data.idc_current.dicom_all`
  ),

  nlst_values_per_series AS (
  SELECT
    # COUNT(DISTINCT(distinct_slice_location_difference_values.SliceLocation_difference)) as num_differences,
    COUNT(distinct_slice_location_difference_values.SliceLocation_difference) as num_differences,
    MAX(ABS(distinct_slice_location_difference_values.SliceLocation_difference)) as max_difference,
    MIN(ABS(distinct_slice_location_difference_values.SliceLocation_difference)) as min_difference,
    ANY_VALUE(nlst_instances_per_series.PatientID) AS PatientID,
    ANY_VALUE(nlst_instances_per_series.StudyInstanceUID) AS StudyInstanceUID,
    distinct_slice_location_difference_values.SeriesInstanceUID AS SeriesInstanceUID,
    ANY_VALUE(nlst_instances_per_series.Modality) AS Modality,
    ANY_VALUE(nlst_instances_per_series.num_instances) AS num_instances,
    ANY_VALUE(nlst_instances_per_series.ImageOrientationPatient) AS ImageOrientationPatient
  FROM
    distinct_slice_location_difference_values
  JOIN
    nlst_instances_per_series
  ON
    nlst_instances_per_series.SeriesInstanceUID = distinct_slice_location_difference_values.SeriesInstanceUID
  WHERE
    nlst_instances_per_series.min_SliceThickness >= 1.5
    AND nlst_instances_per_series.max_SliceThickness <= 3.5
    AND nlst_instances_per_series.num_instances > 100
    AND nlst_instances_per_series.num_instances/nlst_instances_per_series.position_count = 1
    AND nlst_instances_per_series.pixel_spacing_count = 1
    AND nlst_instances_per_series.orientation_count = 1
    AND has_localizer = "false"
    AND ABS(SAFE_CAST(nlst_instances_per_series.ImageOrientationPatient[SAFE_OFFSET(0)] AS float64)) > ABS(SAFE_CAST(nlst_instances_per_series.ImageOrientationPatient[SAFE_OFFSET(1)] AS float64))
    AND ABS(SAFE_CAST(nlst_instances_per_series.ImageOrientationPatient[SAFE_OFFSET(0)] AS float64)) > ABS(SAFE_CAST(nlst_instances_per_series.ImageOrientationPatient[SAFE_OFFSET(2)] AS
```

```sql
float64))
    AND ABS(SAFE_CAST(nlst_instances_per_series.ImageOrientationPatient[SAFE_OFFSET(4)] AS float64)) > ABS(SAFE_CAST(nlst_instances_per_series.ImageOrientationPatient[SAFE_OFFSET(3)] AS float64))
    AND ABS(SAFE_CAST(nlst_instances_per_series.ImageOrientationPatient[SAFE_OFFSET(4)] AS float64)) > ABS(SAFE_CAST(nlst_instances_per_series.ImageOrientationPatient[SAFE_OFFSET(5)] AS float64))
  GROUP BY
    distinct_slice_location_difference_values.SeriesInstanceUID ),

  select_single_series_from_study AS (
  SELECT
    ANY_VALUE(PatientID) AS PatientID,
    StudyInstanceUID,
    ANY_VALUE(SeriesInstanceUID) AS SeriesInstanceUID,
    ANY_VALUE(Modality) AS Modality,
    ANY_VALUE(nlst_values_per_series.num_differences) AS num_differences,
    ANY_VALUE(nlst_values_per_series.max_difference) AS max_difference,
    ANY_VALUE(nlst_values_per_series.min_difference) AS min_difference,
    ANY_VALUE(nlst_values_per_series.num_instances) AS num_instances
  FROM
    nlst_values_per_series
  GROUP BY
    StudyInstanceUID )

  SELECT
    DISTINCT(select_single_series_from_study.SeriesInstanceUID) as SeriesInstanceUID,
    select_single_series_from_study.PatientID,
    select_single_series_from_study.StudyInstanceUID,
    select_single_series_from_study.Modality,
    select_single_series_from_study.num_instances,
    select_single_series_from_study.num_differences,
    select_single_series_from_study.max_difference,
    select_single_series_from_study.min_difference,

CONCAT("https://viewer.imaging.datacommons.cancer.gov/viewer/",select_single_series_from_study.StudyInstanceUID,"?seriesInstanceUID=",select_single_series_from_study.SeriesInstanceUID) AS idc_url,
    dicom_all.gcs_url
```

```
  FROM
    `bigquery-public-data.idc_current.dicom_all` AS dicom_all
  JOIN
    select_single_series_from_study
  ON
    dicom_all.SeriesInstanceUID = select_single_series_from_study.SeriesInstanceUID
  WHERE
    # select_single_series_from_study.num_differences > 1
    # AND
select_single_series_from_study.max_difference/select_single_series_from_study.min_difference >
2
    select_single_series_from_study.num_differences <= 2
    AND
select_single_series_from_study.max_difference/select_single_series_from_study.min_difference <
2
  # FROM
  # select_single_series_from_study
```

Consort diagrams for the cohort selection

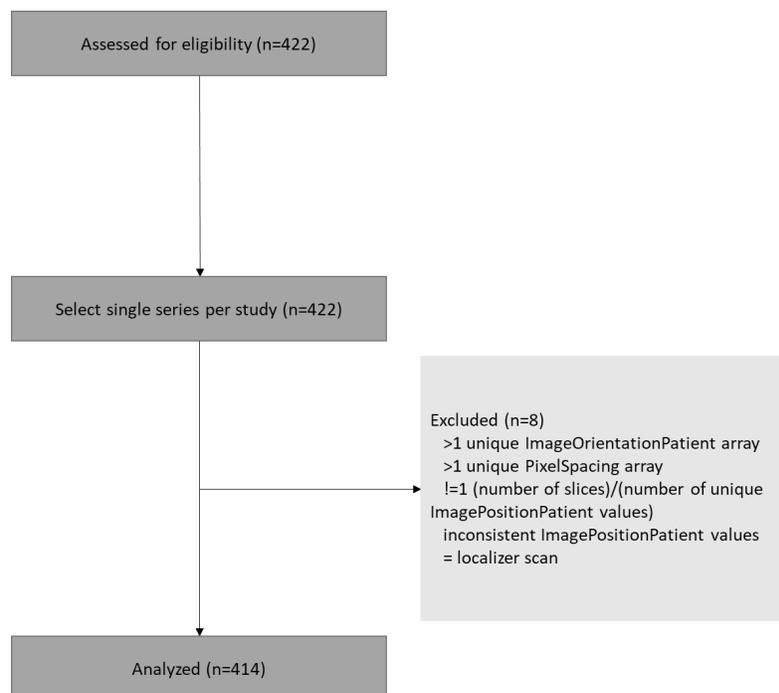

*a) Filtering for the NSCLC-Radiomics collection DICOM CT series selected for the analysis.*

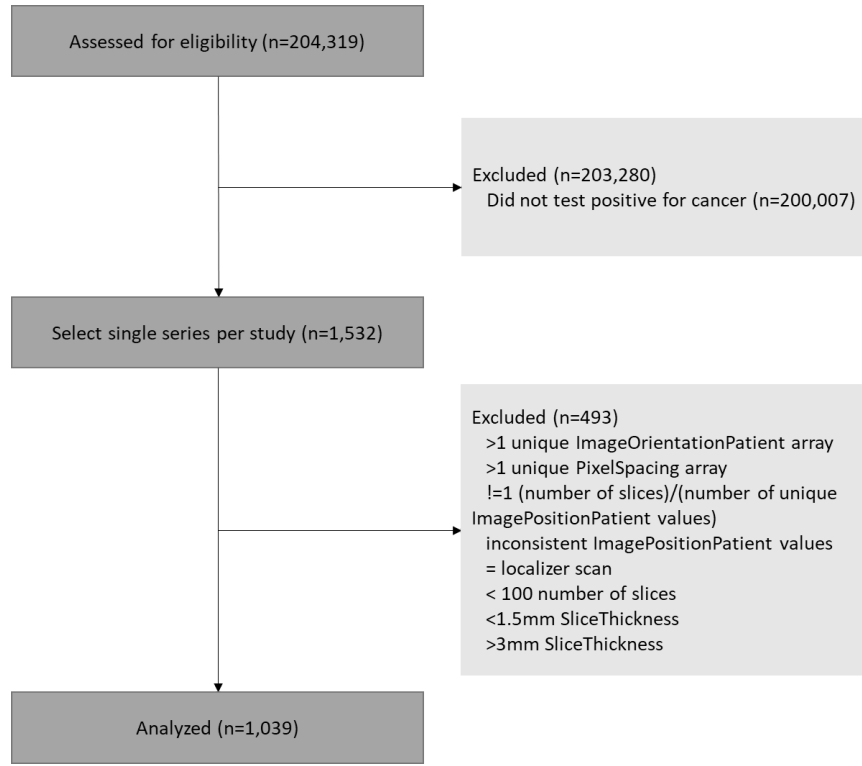

b) Filtering for the NLST collection DICOM CT series selected for the analysis.

*Figure 10 - Overview of the selection process of series for the a) NSCLC-Radiomics collection, and the b) NLST collection.*

## Appendix B

The pyradiomics package[19] (version v3.0.1) was employed to extract 3D shape features from the regions segmented by nnU-Net. The following features were extracted:
1. Elongation
2. Flatness
3. Least Axis Length
4. Major Axis Length
5. Maximum 3D Diameter
6. Mesh Volume
7. Minor Axis Length
8. Sphericity
9. Surface Area
10. Surface Volume Ratio
11. Voxel Volume
12. Compactness 1
13. Compactness 2

14. Spherical Disproportion

# Appendix C

Table 1 below is used for the creation of the json metadata file needed for conversion to a DICOM Segmentation object. The table is also available here [https://github.com/ImagingDataCommons/nnU-Net-BPR-annotations/blob/main/nnunet/data/nnunet_segments_code_mapping.csv].

| Segment | Finding CodingScheme Designator | Finding CodeValue | Finding CodeMeaning | FindingSite CodingScheme Designator | FindingSite CodeValue | FindingSite CodeMeaning |
|---|---|---|---|---|---|---|
| Esophagus | SCT | 113343008 | Organ | SCT | 32849002 | Esophagus |
| Heart | SCT | 113343008 | Organ | SCT | 80891009 | Heart |
| Trachea | SCT | 113343008 | Organ | SCT | 44567001 | Trachea |
| Aorta | SCT | 113343008 | Organ | SCT | 15825003 | Aorta |

Table 1 - The segments code mapping file used to create json metadata file for conversion of NifTi files to DICOM Segmentation objects

# Appendix D

Table 2 below was used for the creation of the DICOM Structured Reports for holding radiomics features. The table is also available here [https://github.com/ImagingDataCommons/nnU-Net-BPR-annotations/blob/main/nnunet/data/nnunet_shape_features_code_mapping.csv]. 14 3D shape features were chosen for the analysis, where each feature corresponds to the IBSI standard[26], and is described by the quantity and units.

| shape_feature | quantity_ CodingSchemeDesignator | quantity_ CodeValue | quantity_ CodeMeaning | units_ CodingSchemeDesignator | units_ CodeValue | units_ CodeMeaning |
|---|---|---|---|---|---|---|
| Elongation | IBSI | Q3CK | Elongation | UCUM | mm | millimeter |
| Flatness | IBSI | N17B | Flatness | UCUM | mm | millimeter |
| LeastAxisLength | IBSI | 7J51 | Least Axis in 3D Length | UCUM | mm | millimeter |
| MajorAxisLength | IBSI | TDIC | Major Axis in 3D Length | UCUM | mm | millimeter |
| Maximum3DDiameter | IBSI | L0JK | Maximum 3D Diameter of a Mesh | UCUM | mm | millimeter |
| MeshVolume | IBSI | RNU0 | Volume of Mesh | UCUM | mm3 | cubic millimeter |
| MinorAxisLength | IBSI | P9VJ | Minor Axis in 3D Length | UCUM | mm | millimeter |
| Sphericity | IBSI | QCFX | Sphericity | UCUM | 1 | no units |
| SurfaceArea | IBSI | C0JK | Surface Area of Mesh | UCUM | mm2 | square millimeter |
| SurfaceVolumeRatio | IBSI | 2PR5 | Surface to Volume Ratio | UCUM | /mm | per millimeter |
| VoxelVolume | IBSI | YEKZ | Volume from Voxel Summation | UCUM | mm3 | cubic millimeter |
| Compactness1 | IBSI | SKGS | Compactness 1 | UCUM | 1 | no units |
| Compactness2 | IBSI | BQWJ | Compactness 2 | UCUM | 1 | no units |
| SphericalDisproportion | IBSI | KRCK | Spherical Disproportion | UCUM | 1 | no units |

Table 2 - The shape features code mapping file used to create the TID1500 DICOM Structured Reports

# Appendix E

The landmarks produced from the body part prediction neural network were saved as DICOM Structured Reports, where a single report is saved for each series analyzed. The highdicom package[27] was used to generate the SR. Table 3 below holds the information concerning the mapping of each of the body part regression codes to specific bones and landmarks. The optional modifier holds information about the location of the particular anatomical landmark. The table is also available here [https://github.com/ImagingDataCommons/nnU-Net-BPR-annotations/blob/main/bpr/data/bpr_landmarks_code_mapping.csv]. The list of landmarks was retrieved from the landmark look up table.[8]

| BPR_code | CodingSchemeDesignator | CodeValue | CodeMeaning | modifier_CodingSchemeDesignator | modifier_CodeValue | modifier_CodeMeaning |
|---|---|---|---|---|---|---|
| | | | | | | |

| C1 | SCT | 14806007 | C1 vertebra | DCM | 111010 | Center |
| --- | --- | --- | --- | --- | --- | --- |
| C2 | SCT | 39976000 | C2 vertebra | DCM | 111010 | Center |
| C3 | SCT | 113205007 | C3 vertebra | DCM | 111010 | Center |
| C4 | SCT | 5329002 | C4 vertebra | DCM | 111010 | Center |
| C5 | SCT | 36978003 | C5 vertebra | DCM | 111010 | Center |
| C6 | SCT | 36054005 | C6 vertebra | DCM | 111010 | Center |
| C7 | SCT | 87391001 | C7 vertebra | DCM | 111010 | Center |
| Th1 | SCT | 64864005 | T1 vertebra | DCM | 111010 | Center |
| Th2 | SCT | 53733008 | T2 vertebra | DCM | 111010 | Center |
| Th3 | SCT | 1626008 | T3 vertebra | DCM | 111010 | Center |
| Th4 | SCT | 73071006 | T4 vertebra | DCM | 111010 | Center |
| Th5 | SCT | 56401006 | T5 vertebra | DCM | 111010 | Center |
| Th6 | SCT | 45296009 | T6 vertebra | DCM | 111010 | Center |
| Th7 | SCT | 62487009 | T7 vertebra | DCM | 111010 | Center |
| Th8 | SCT | 11068009 | T8 vertebra | DCM | 111010 | Center |
| Th9 | SCT | 82687006 | T9 vertebra | DCM | 111010 | Center |
| Th10 | SCT | 7610001 | T10 vertebra | DCM | 111010 | Center |
| Th11 | SCT | 12989004 | T11 vertebra | DCM | 111010 | Center |
| Th12 | SCT | 23215003 | T12 vertebra | DCM | 111010 | Center |
| L1 | SCT | 66794005 | L1 vertebra | DCM | 111010 | Center |
| L2 | SCT | 14293000 | L2 vertebra | DCM | 111010 | Center |
| L3 | SCT | 36470004 | L3 vertebra | DCM | 111010 | Center |
| L4 | SCT | 11994002 | L4 vertebra | DCM | 111010 | Center |
| L5 | SCT | 49668003 | L5 vertebra | DCM | 111010 | Center |
| pelvis_start | SCT | 8106092008 | Pelvis | SCT | 42161009 | Bottom |
| femur_end | SCT | 71341001 | Femur | SCT | 421812003 | Top |
| pelvis_end | SCT | 8106092008 | Pelvis | SCT | 421812003 | Top |
| kidney | SCT | 64033007 | Kidney | SCT | 42161009 | Bottom |
| lung_start | SCT | 39607008 | Lung | SCT | 42161009 | Bottom |
| lung_end | SCT | 39607008 | Lung | SCT | 421812003 | Top |
| liver_end | SCT | 10200004 | Liver | SCT | 421812003 | Top |
| teeth | SCT | 28035005 | Teeth, gums and supporting structures | DCM | 111010 | Center |
| nose | SCT | 45206002 | Nose | SCT | 42161009 | Bottom |
| eyes_end | SCT | 81745001 | Eye | SCT | 421812003 | Top |
| head_end | SCT | 69536005 | Head | SCT | 421812003 | Top |

Table 3 - The landmark code mapping file used to create the TID1500 DICOM Structured Reports

## Appendix F

Inference from the body part prediction neural network also includes a region(s) assignment for each axial slice. This region assignment information was saved as a DICOM Structured Report, where a single report is saved for each series analyzed. The highdicom package[27] was used to generate the SR. The table is also available here [https://github.com/ImagingDataCommons/nnU-Net-BPR-annotations/blob/main/bpr/data/bpr_regions_code_mapping.csv].

| BPR_code_region | CodingSchemeDesignator | CodeValue | CodeMeaning |
|---|---|---|---|
| legs | SCT | 30021000 | Legs |
| pelvis | SCT | 12921003 | Pelvis |
| abdomen | SCT | 113345001 | Abdomen |
| chest | SCT | 51185008 | Chest |
| shoulder-neck | SCT | 45048000 | Neck |

| head | SCT | 69536005 | Head |

Table 4 - The region code mapping file used to create the TID1500 DICOM Structured Reports

# References

1.  Fedorov, A. *et al.* NCI Imaging Data Commons. *Cancer Res.* **81**, 4188–4193 (2021).

2.  Clark, K. *et al.* The Cancer Imaging Archive (TCIA): maintaining and operating a public information repository. *J. Digit. Imaging* **26**, 1045–1057 (2013).

3.  Aerts, H., Velazquez, E. R., Leijenaar, R. T. & Parmar, C. Data from NSCLC-radiomics. *The cancer imaging archive*.

4.  Aerts, H. J. W. L. *et al.* Decoding tumour phenotype by noninvasive imaging using a quantitative radiomics approach. *Nat. Commun.* **5**, 4006 (2014).

5.  Clark, K. National Lung Screening Trial. https://wiki.cancerimagingarchive.net/display/NLST/National+Lung+Screening+Trial.

6.  Team, T. N. L. S. T. R. & The National Lung Screening Trial Research Team. Reduced Lung-Cancer Mortality with Low-Dose Computed Tomographic Screening. *New England Journal of Medicine* vol. 365 395–409 Preprint at https://doi.org/10.1056/nejmoa1102873 (2011).

7.  Isensee, F., Jaeger, P. F., Kohl, S. A. A., Petersen, J. & Maier-Hein, K. H. nnU-Net: a self-configuring method for deep learning-based biomedical image segmentation. *Nat. Methods* **18**, 203–211 (2021).

8.  Schuhegger, S. Body Part Regression for CT Images. *arXiv [eess.IV]* (2021).

9.  Wilkinson, M. D. *et al.* The FAIR Guiding Principles for scientific data management and stewardship. *Sci Data* **3**, 160018 (2016).

10. 1 scope and field of application.

https://dicom.nema.org/medical/dicom/current/output/chtml/part01/chapter_1.html.

11. *dicomsort: DICOM sorting utility*. (Github).

12. *dcm2niix: dcm2nii DICOM to NIfTI converter: compiled versions available from NITRC*. (Github).

13. Antonelli, M. *et al.* The Medical Segmentation Decathlon. *Nat. Commun.* **13**, 4128 (2022).

14. Ji, Y. *et al.* Amos: A large-scale abdominal multi-organ benchmark for versatile medical image segmentation. *Adv. Neural Inf. Process. Syst.* **35**, 36722–36732 (2022).

15. Isensee, F. *et al.* nnU-Net: Self-adapting Framework for U-Net-Based Medical Image Segmentation. *arXiv [cs.CV]* (2018).

16. Lambert, Z., Petitjean, C., Dubray, B. & Kuan, S. SegTHOR: Segmentation of Thoracic Organs at Risk in CT images. in *2020 Tenth International Conference on Image Processing Theory, Tools and Applications (IPTA)* 1–6 (2020). doi:10.1109/IPTA50016.2020.9286453.

17. *nnUNet*. (Github).

18. Krishnaswamy, D., Bontempi, D., Clunie, D., Aerts, H. & Fedorov, A. *AI-derived annotations for the NLST and NSCLC-Radiomics computed tomography imaging collections*. (2023). doi:10.5281/zenodo.7822904.

19. van Griethuysen, J. J. M. *et al.* Computational Radiomics System to Decode the Radiographic Phenotype. *Cancer Res.* **77**, e104–e107 (2017).

20. *BodyPartRegression*. (Github).

21. Ziegler, E. *et al.* Open Health Imaging Foundation Viewer: An Extensible Open-Source Framework for Building Web-Based Imaging Applications to Support Cancer Research. *JCO Clin Cancer Inform* **4**, 336–345 (2020).

22. Zeleznik, R. *et al.* Deep convolutional neural networks to predict cardiovascular risk from computed tomography. *Nat. Commun.* **12**, 715 (2021).

23. Gierada, D. S. *et al.* Quantitative CT assessment of emphysema and airways in relation to lung cancer risk. *Radiology* **261**, 950–959 (2011).

24. IDC. https://portal.imaging.datacommons.cancer.gov/.

25. Herz, C. *et al.* dcmqi: An Open Source Library for Standardized Communication of Quantitative Image Analysis Results Using DICOM. *Cancer Res.* **77**, e87–e90 (2017).

26. Zwanenburg, A. *et al.* The Image Biomarker Standardization Initiative: Standardized Quantitative Radiomics for High-Throughput Image-based Phenotyping. *Radiology* vol. 295 328–338 Preprint at https://doi.org/10.1148/radiol.2020191145 (2020).

27. Bridge, C. P. *et al.* Highdicom: a Python Library for Standardized Encoding of Image Annotations and Machine Learning Model Outputs in Pathology and Radiology. *J. Digit. Imaging* **35**, 1719–1737 (2022).

28. Clunie, D. A. dicom3tools. *Dicom3tools Software* https://www.dclunie.com/dicom3tools.html (2022).

29. Clunie, D. A. PixelMed Java DICOM Toolkit. *PixelMed Publishing* http://www.pixelmed.com/index.html#PixelMedJavaDICOMToolkit (2017).

30. Doherty, D. Pediatric Critical Care – Fourth Edition. *Canadian Journal of Anesthesia/Journal canadien d'anesthésie* **59**, 427–428 (2012).

31. Koitka, S. *pydicom-seg: Python package for DICOM-SEG medical segmentation file reading and writing*. (Github).

32. *itkwidgets: Interactive Jupyter widgets to visualize images, point sets, and meshes in 2D*

*and 3D*. (Github).